\documentclass[10pt, a4paper]{article}

\usepackage{lrec2026} 
\usepackage{xcolor}
\usepackage{pifont}
\usepackage{graphicx}
\usepackage{booktabs}
\usepackage{enumitem}
\usepackage{ulem}
\usepackage{todonotes}
\usepackage{multirow}
\usepackage{float}

\title{HateMirage: An Explainable Multi-Dimensional Dataset for Decoding Faux Hate and Subtle Online Abuse}

\name{Sai Kartheek Reddy Kasu$^{1\ast}$\thanks{$^\ast$Work done at FLaMe.nlp Lab, IIIT Delhi.}, 
Shankar Biradar$^{2}$, 
Sunil Saumya$^{1}$, 
Md. Shad Akhtar$^{3}$}

\address{$^{1}$Indian Institute of Information Technology Dharwad \\
         $^{2}$Manipal Institute of Technology, Manipal Academy of Higher Education
Manipal, Karnataka, India \\
         $^{3}$Indraprastha Institute of Information Technology Delhi \\
         saikartheekreddykasu@gmail.com, 
         shankar.biradar@manipal.edu,\\
         sunil.saumya@iiitdwd.ac.in, shad.akhtar@iiitd.ac.in}
\abstract{
Subtle and indirect hate speech remains an underexplored challenge in online safety research, particularly when harmful intent is embedded within misleading or manipulative narratives. Existing hate speech datasets primarily capture overt toxicity, underrepresenting the nuanced ways misinformation can incite or normalize hate. To address this gap, we present \textit{HateMirage}, a novel dataset of Faux Hate comments designed to advance reasoning and explainability research on hate emerging from fake or distorted narratives. The dataset was constructed by identifying widely debunked misinformation claims from fact-checking sources and tracing related YouTube discussions, resulting in 4,530 user comments. Each comment is annotated along three interpretable dimensions: Target (who is affected), Intent (the underlying motivation or goal behind the comment), and Implication (its potential social impact). Unlike prior explainability datasets such as HateXplain and HARE, which offer token-level or single-dimensional reasoning, HateMirage introduces a multi-dimensional explanation framework that captures the interplay between misinformation, harm, and social consequence. We benchmark multiple open-source language models on HateMirage using ROUGE-L F1 and Sentence-BERT similarity to assess explanation coherence. Results suggest that explanation quality may depend more on pretraining diversity and reasoning-oriented data rather than on model scale alone. By coupling misinformation reasoning with harm attribution, HateMirage establishes a new benchmark for interpretable hate detection and responsible AI research.\\ 
\newline \Keywords{Faux Hate; Hate Speech; Misinformation; Explainable NLP; Social Media Safety; Benchmark Dataset}}

\begin{document}
\maketitleabstract

\section{Introduction}
\textcolor{red}{This paper contains examples of faux hate comments, included solely for research purposes. Some readers may find the content offensive or distressing. Reader discretion is advised.}

\noindent Online platforms such as YouTube, Facebook, and X (formerly Twitter) have become central spaces for communication and opinion exchange but also amplify harmful content, including hate speech and misinformation \cite{schmidt2017survey, fortuna2018survey}. Despite advances in moderation, subtle and indirect hate, where hostility is masked through humor, irony, or fabricated narratives, remains difficult to detect. Such content often avoids explicit slurs or profanity, making it ambiguous even for human annotators and challenging for automated systems.

Existing hate speech datasets, including HateXplain \cite{mathew2021hatexplain}, Hostility Detection \cite{bhardwaj2020hostility}, studies on geographic and cultural bias \cite{tonneau2024languages}, and SemEval series corpora \cite{zampieri2020semeval, premjith2024findings}, have enabled progress in classification and target identification. However, they largely focus on overt hate and surface-level toxicity cues. This limits model interpretability and fails to capture forms of hate that are manufactured, fueled, or triggered by misinformation.

Building on the challenges of detecting subtle, context-dependent harm, recent work has identified \textit{Faux Hate} \cite{biradar2025faux}: hate speech expressed indirectly through misinformation or deceptive narratives. Unlike traditional hate, which uses overt hostility, Faux Hate embeds prejudice within misleading claims, making its harmful intent implicit. For example, a comment may falsely accuse a community of spreading disease, conveying resentment while hiding the premise’s falsity. Detecting such content requires contextual reasoning and background knowledge, which makes automated detection difficult. For instance, \autoref{tab:hate_vs_fauxhate_comments} contrasts traditional hatespeeches, which relies on direct attacks, with Faux Hate, which embeds prejudice in misleading or fabricated narratives, highlighting the need for contextual reasoning to interpret it effectively. 

\begin{table}[!t]
\begin{center}
\resizebox{\columnwidth}{!}{%
\begin{tabular}{p{8cm}}
\toprule
\textbf{Typical Hateful Comment} \\ \toprule
-- I dislike the skin colour of people from \texttt{$<$country$>$}. \\
-- \texttt{$<$country$>$} is dishonest toward everyone; we should boycott it. \\
-- \texttt{$<$group/religion$>$} people are ruining our society; they should leave. \\
\midrule
\textbf{Faux Hate Comments} \\
\midrule
-- \texttt{$<$country$>$} is spreading COVID on purpose to harm the world. \\
-- It’s not love, it’s a plan; \texttt{$<$group/religion$>$} use relationships to change people.\\
-- The virus came from \texttt{$<$country$>$} labs; they engineered it to destroy us. \\
\bottomrule
\end{tabular}%
}
\caption{Illustrative examples comparing Traditional Hate comments with Faux Hate comments. Placeholders such as \texttt{COUNTRY} and \texttt{GROUP/RELIGION} denote target entities abstractly.}
\label{tab:hate_vs_fauxhate_comments}
\end{center}
\end{table}

Prior work \cite{biradar2024proceedings, biradar2025faux} mainly focused on the classification of Faux Hate, leaving the task of explaining its reasoning and potential harm largely unexplored. This limitation hinders a deeper understanding of how misinformation shapes and triggers hate. To address this gap, we introduce \textit{HateMirage}\footnote{The term reflects the illusory nature of Faux Hate, where hateful expression appears genuine but is in fact shaped or amplified by fake narratives.}, a dataset for reasoning and explainability in misinformation-driven hate. It contains 4,530 comments collected using a two-step process: we first gathered widely debunked claims from fact-checking websites (e.g., \textit{AltNews}\footnote{https://www.altnews.in/}, \textit{FactChecker}\footnote{https://www.factchecker.in/}), then identified YouTube videos related to these claims and extracted associated comments. Each comment is annotated with Target, Intent, and Implication, capturing the reasoning that links deceptive narratives to harm. 


We benchmarked several open-source language models ranging from 1B to 8B parameters for the structured explanation generation task. The HateMirage dataset serves as a unified testbed, providing a consistent setup to evaluate model reasoning across the Target, Intent, and Implication dimensions. Each model was evaluated in two settings: zero-shot and RAG-based. Performance was measured using Sentence-BERT similarity and ROUGE-L F1 to assess semantic and lexical coherence. The analysis highlights variations across model architectures and settings, reflecting the complexity of reasoning required for multi-dimensional explanation generation.

HateMirage thus makes implicit harm explicit, supporting research on interpretable NLP, explanation generation, and context-aware moderation \cite{mathew2021hatexplain, yang2023hare}. By exposing the logic connecting misinformation to prejudice, it enables models and analysts to justify predictions, which is critical for trustworthy content moderation.

\paragraph{Contributions:} The key contributions of this work are summarized as follows:

\begin{itemize}[leftmargin=*, noitemsep]
    \item \textbf{HateMirage Dataset:} We present \textit{HateMirage}, a dataset of 4,530 Faux Hate comments, annotated with multi-dimensional explanations across Target, Intent, and Implication. These annotations capture the implicit narratives and reasoning underlying misinformation-driven hate speech, enabling interpretable analysis beyond surface-level classification. 
    \item \textbf{Model Benchmarking:} We evaluate a range of small and large language models on the task of explanation generation for Faux Hate, establishing initial baselines and highlighting the challenges of structured reasoning in this domain.
    \item \textbf{Research Testbed for Explainable NLP:} By explicitly linking misinformation cues to hateful discourse, HateMirage provides a valuable resource for studying reasoning, interpretability, and responsible AI in online communities.
\end{itemize}

\paragraph{Reproducibility:} Prompt templates, annotation guidelines, RAG configuration, source code, and supporting fabricated claims are publicly available at \url{https://github.com/Sai-Kartheek-Reddy/HateMirage}. The HateMirage dataset is available for academic and research purposes only upon acceptance of the data usage agreement (\href{https://forms.gle/jRW7jKf5ASnbb8ia9}{link}).

\section{Literature Review}\label{literature}

Current literature on online harmful content has largely treated hate speech and misinformation as separate domains, focusing on developing models and resources to detect either hate or fake content individually. However, the intersection of these two phenomena, where users propagate hateful comments in the context of widely debunked claims, remains largely underexplored \cite{biradar2025faux, mosleh2024misinformation}. Understanding this intersection is crucial, as it captures nuanced forms of social media manipulation and the complex interplay between intent, target, and potential consequences. The HateMirage dataset is designed to address this gap by providing multi-dimensional annotations that reflect both hate and misinformation in a unified framework.

\begin{table}[!t]
\begin{center}
\resizebox{\columnwidth}{!}{%
\begin{tabular}{l c c c c}
\toprule
Dataset & Hate & Fake & Explanation & Interdisciplinary \\
\toprule
HateXplain \cite{mathew2021hatexplain} & \ding{51} & \textcolor{red}{\ding{55}} & \textcolor{red}{\ding{55}} & \textcolor{red}{\ding{55}} \\
HatEval \cite{basile2019semeval} & \ding{51} & \textcolor{red}{\ding{55}} & \textcolor{red}{\ding{55}} & \textcolor{red}{\ding{55}} \\ 
Implicit Hate \cite{elsherief-etal-2021-latent} & \ding{51} & \textcolor{red}{\ding{55}} & \ding{51} & \textcolor{red}{\ding{55}} \\

HARE \cite{yang2023hare} & \ding{51} & \textcolor{red}{\ding{55}} & \ding{51} & \textcolor{red}{\ding{55}} \\

CoAID \cite{cui2020coaid} & \textcolor{red}{\ding{55}} & \ding{51} & \textcolor{red}{\ding{55}} & \textcolor{red}{\ding{55}} \\

FNVE \cite{chen2025multimodal} & \textcolor{red}{\ding{55}} & \ding{51} & \ding{51} & \textcolor{red}{\ding{55}} \\

Hostile \cite{bhardwaj2020hostility} & \ding{51} & \ding{51} & \textcolor{red}{\ding{55}} & \textcolor{red}{\ding{55}} \\

Deceptive Humor \cite{kasu2025deceptive} & \ding{51} & \ding{51} & \textcolor{red}{\ding{55}} & \ding{51} \\
\midrule
HateMirage (Proposed) & \ding{51} & \ding{51} & \ding{51} & \ding{51} \\
\bottomrule
\end{tabular}%
}
\caption{Comparison of existing datasets across four key dimensions: hate speech labeling, fake/misinformation labeling, structured/multi-dimensional explanatory annotations, and interdisciplinary scope. The interdisciplinary attribute denotes the integration of two distinct domains within a single dataset, allowing their interaction to be studied rather than treated separately. \ding{51} indicates the presence of a feature, while \textcolor{red}{\ding{55}} indicates its absence. Unlike prior resources, HateMirage incorporates all four dimensions, supporting richer analysis and interpretability.}
\label{tab:dataset_comparison}
\end{center}
\end{table}

\paragraph{Hate Speech:} While several prior datasets have contributed significantly to hate speech and misinformation research, they remain limited in scope with respect to structured explanatory annotations as shown in \autoref{tab:dataset_comparison}. HateXplain \cite{mathew2021hatexplain} focuses purely on hate speech and includes token-level rationales highlighting which parts of a comment are hateful, yet it does not explain why such comments are made or what intent and implication they convey. HatEval \cite{basile2019semeval}, another benchmark dataset, provides hate labels for multilingual data but lacks any explanatory or interpretive layer, restricting analysis to surface-level classification. Implicit Hate \cite{elsherief-etal-2021-latent} advances the field by capturing latent and indirect hate through contextual reasoning, but it still does not link hate expression to misinformation cues or structured components like target or intent. Similarly, HARE \cite{yang2023hare} offers explanatory information for hate categories but remains confined to hateful content without incorporating misinformation or interdisciplinary context.

\paragraph{Misinformation:} On the misinformation side, CoAID \cite{cui2020coaid} and FNVE \cite{chen2025multimodal} provide extensive fake news data, with the latter including multimodal evidence and explanatory text for fact-checking. However, both lack any representation of hate or harmful intent, making them unsuitable for studying hate arising from false narratives. Hostile \cite{bhardwaj2020hostility} combines hate and fake content but treats them as independent categories; the hate content is not causally derived from misinformation and vice versa. Deceptive Humor \cite{kasu2025deceptive}, though conceptually closer to Faux Hate, presents misinformation veiled as humor, which complicates detection and amplifies misinformation propagation along with harmful reactions, yet it too omits structured explanatory dimensions.

In contrast, HateMirage deals with faux hate comments (i.e, fake = 1 and hate = 1 ) with multi-dimensional structured explanations spanning target, intent, and implication. This comprehensive design enables deeper analysis of how misinformation shapes hateful discourse and supports interpretability across social, psychological, and linguistic perspectives.


\section{Dataset Overview} \label{dataset_overview}
In this section, we present a comprehensive overview of the HateMirage dataset. We describe the process of collecting widely debunked fake claims from reputable fact-checking websites, followed by scraping user comments from international English news channels. We explain the automated labeling procedure for identifying fake and hate comments, and detail the generation of structured explanations capturing the multi-dimensional aspects of each comment, including target, intent, and implication. Finally, we provide quantitative statistics and representative examples to illustrate the linguistic complexity and diversity of the dataset.

\subsection{Collection of Debunked Fake Claims}
The foundation of the HateMirage dataset is a curated set of widely debunked fake claims. These claims were sourced from reputable fact-checking websites such as AltNews\footnote{\url{https://www.altnews.in/}} and FactCheckers\footnote{\url{https://www.factchecker.in/}}, which maintain comprehensive repositories of misinformation that has been verified and refuted. We focused on claims that were widely circulated on social media to ensure the dataset captures content that is both impactful and relevant. Each claim was carefully reviewed to confirm its status as false or misleading, ensuring that subsequent data collection and analysis are grounded in verified information. This curated set of claims serves as the starting point for gathering user comments and generating structured explanations, providing a reliable basis for studying the intersection of misinformation and hate speech.

\subsection{Data Collection and Labeling}
We collect user comments from the YouTube sections of reputable international English news channels, capturing both purely English comments and code-mixed Hindi-English comments to reflect the linguistic diversity of social media discourse. International channels were chosen to reduce potential bias that may arise from local news coverage favoring one side. Each comment is labeled with two binary annotations: Hate (0/1) and Fake (0/1). The labeling process is automated using a GPT-4 model \cite{achiam2023gpt}, which is augmented with Retrieval-Augmented Generation (RAG) \cite{lewis2020retrieval} to provide additional grounded context for more accurate predictions. To maintain high data quality, we incorporate a human-in-the-loop validation stage even in the presence of RAG. While retrieval provides supporting context, model outputs can still suffer from hallucinations, cultural misinterpretations, or sensitivity to code-mixed language. Therefore, randomly sampled annotations are manually reviewed by human evaluators to verify label correctness, resolve ambiguous cases, and detect systematic biases. This process acts as a quality control mechanism, increasing confidence in the reliability of the automatically generated labels and ensuring that errors do not propagate into the final dataset.

\subsection{Structured Explanation Generation}
To better understand the underlying dynamics of fake and hateful comments, we first curated only the \textit{Faux Hate} instances, i.e., comments labeled as both \textit{fake = 1} and \textit{hate = 1}. These instances were drawn from both the existing FEUD dataset \cite{biradar2025faux}, previously collected from Twitter and YouTube, as well as additional comments newly collected from YouTube channels. The collected Hindi-English code-mixed comments were then translated into English using the GPT-4 model, leveraging its extensive pre-trained multilingual understanding. We intentionally avoided text pre-processing operations such as punctuation or special character removal to preserve the raw linguistic style and contextual essence typical of social media discourse. English translation was preferred since most LLMs demonstrate stronger performance and broader support for English text, ensuring consistency in downstream processing and evaluation.

Building on this curated dataset, we aim to move beyond binary classification and delve into the underlying mechanisms of hate expression. To achieve this, we generate structured explanations that capture three key dimensions: the \textit{Target} of the comment, the \textit{Intent} behind it, and its potential \textit{Implications} on social discourse. These explanations offer a multi-dimensional understanding of faux hate behavior, enabling a richer analysis of user motivations, the nature of attacks, and their societal impact.

We employ GPT-4 to generate structured explanations due to its extensive pretrained world knowledge, strong reasoning abilities, and robust language understanding. Faux hate comments often embed misinformation implicitly and rely on nuanced socio-political or cultural references, which makes simple pattern matching or smaller models insufficient. GPT-4 can effectively infer unstated targets, identify underlying intent, and articulate broader implications without task-specific fine-tuning. To ensure factual grounding, we adopt a RAG approach that supplements GPT-4 with relevant fact-checked evidence during generation. 

\paragraph{Quality Validation (Human-in-the-Loop):} The explanation generation process includes a human-in-the-loop component, where sampled outputs are continuously monitored and verified during generation to maintain quality, reliability, and consistency. Quality is evaluated across three objective dimensions: factual consistency, ensuring that each explanation aligns with verified evidence without introducing unsupported claims; semantic coherence, requiring logical alignment between the comment and its Target, Intent, and Implication fields; and linguistic clarity, emphasizing grammatical correctness, conciseness, and interpretability. We discard the samples that do not meet the above quality criteria. 

To operationalize this workflow, two reviewers (one PhD scholar specializing in hate speech and one graduate student) participated in validation. As the dataset was generated in batches, we randomly sampled 10–15\% of instances from each batch for manual inspection. Reviewers assessed explanations for factual consistency, semantic coherence, and linguistic clarity against the retrieved evidence. When inconsistencies were identified, we refined the retrieval pool and regenerated the outputs. Disagreements were resolved through discussion, and explanations were manually consolidated when necessary. All reviewers provided informed consent. This iterative auditing process helped prevent systematic errors and improved retrieval coverage and explanation reliability.

\subsection{Representative Examples} 
\begin{table*}[!ht]
\begin{center}
\resizebox{0.98\textwidth}{!}{%
\begin{tabular}{p{3.8cm} p{3.8cm} p{2cm} p{4.2cm} p{4.2cm}}
\toprule
\textbf{Comment} & \textbf{Fake Claim Base} & \textbf{Target} & \textbf{Intent} & \textbf{Implication} \\
\toprule
\textit{The coronavirus is just a scam planned by [Political Party] and [Country] to fool us all.} 
& "[Country] is Using Covid as a Biological Weapon"
& [Political Party], [Country]
& To link [Political Party] and [Country] to a conspiracy theory about COVID-19.
& Could fuel public anger against [Political Party] and [Country]. \\

\midrule

\textit{[Religious Group] and [Political Party], with [Person] in the mix, spread COVID to ruin [City] and harm [Country].}
& "[Religious Group] and [Political Party] collaborated to spread COVID-19."
& [Political Party], [Person], [Religious Group]
& To accuse [Political Party], [Person], and [Religious Group] of intentionally worsening the pandemic.
& Could stir up communal tensions and increase distrust in political and religious groups. \\

\midrule

\textit{[Religious Group] men are trapping innocent [Community] women in fake love to convert them. This is how [Country] is being taken over slowly.}
& "Love Jihad: [Religious Group] men are systematically luring women from other communities to convert them."
& [Religious Group]
& To portray [Religious Group] as deceptive and predatory, reinforcing the fabricated 'Love Jihad' narrative.
& Could deepen mistrust between communities, normalize religious profiling, and justify hostility toward [Religious Group]. \\

\bottomrule
\end{tabular}%
}
\caption{Representative examples of Faux Hate comments illustrating how explanations are derived from comments rooted in fake or manipulative narratives. Each example decomposes the comment into its underlying fake claim base, the entities targeted, inferred author intent, and the broader social or emotional implications.}
\label{tab:representative_examples}
\end{center}
\end{table*}

The representative examples in \autoref{tab:representative_examples} illustrate the multi-dimensional reasoning framework underlying HateMirage explanations. Each comment is deconstructed along three critical axes: \textit{Target}, \textit{Intent}, and \textit{Implication}, which together delineate not just who is attacked, but why and with what social consequence. Identifying the \textit{Target} demands recognizing both explicit mentions and implicit associations embedded within misinformation. Inferring the \textit{Intent} requires discerning the author’s underlying motive, whether to vilify, delegitimize, or polarize, often veiled beneath seemingly informational or conspiratorial tones. Finally, articulating the \textit{Implication} captures the downstream effects on collective perception, social cohesion, or intergroup trust, bridging linguistic form and societal impact. Capturing these dimensions is inherently challenging, as they often co-occur in subtle, context-dependent ways, requiring interpretive reasoning beyond surface-level toxicity cues. Together, these structured explanations provide a granular view of how deceptive narratives operationalize hate, advancing both interpretability and accountability in computational social science.



\subsection{Quantitative Analysis} \autoref{tab:quant_summary} presents a quantitative summary of the HateMirage dataset across all fields: Comments, Target, Intent, and Implication. The dataset contains linguistically diverse user comments, ranging from short statements to lengthy posts (up to 800 tokens). The table reports key metrics for each field: Avg Tokens (average number of tokens per text), Median Tokens (median number of tokens per text), Min/Max Tokens (minimum and maximum tokens per text), Avg Sentence Length (average number of tokens per sentence), Max Sentence Length (longest sentence in tokens), Multi-Sentence \% (percentage of texts containing more than one sentence), and Total Tokens (overall token count across all texts). Structured explanations are concise, with most Target annotations consisting of one or two tokens, while Intent and Implication texts typically span 14-15 tokens on average. Over 75\% of Comments contain multiple sentences, highlighting the linguistic complexity present in user-generated content. These statistics illustrate the richness and diversity of the dataset, supporting both hate/fake classification and multi-dimensional explanatory analysis.

\begin{table*}[!ht]
\begin{center}
\resizebox{0.95\textwidth}{!}{%
\begin{tabular}{l l r r r r r r r r }
\toprule
& \bf Field & \bf Avg Tokens & \bf Median Tokens & \bf Min Tokens & \bf Max Tokens & \bf Avg Sent & \bf Max Sent Len &  \bf Multi-Sent(\%) & \bf Total Tokens \\ \toprule

\multirow{4}{*}{\bf Train (3,624 samples)} 
& Comments    & 30.56 & 29 & 3 & 800 & 2.82 & 240 & 76.49 & 110741 \\
& Target      & 1.48  & 1 & 1 & 27 & 1.00 & 12 & 0.30  & 5338\\
& Intent      & 14.66 & 14 & 5 & 39 & 1.02 & 34 & 1.41  & 53101 \\
& Implication & 15.16 & 15 & 6 & 48 & 1.01 & 38 & 0.86  & 54878 \\
\midrule
\multirow{4}{*}{\bf Test (906 samples)} 
& Comments    & 30.28 & 27 & 4 & 316 & 2.84 & 78 & 73.18 & 27438 \\
& Target      & 1.51  & 1 & 1 & 44 & 1.01 & 24 & 0.44  & 1371\\
& Intent      & 14.57 & 14 & 5 & 40 & 1.02 & 33 & 1.43  & 13204 \\
& Implication & 15.14 & 14 & 5 & 40 & 1.01 & 40 & 0.66  & 13706 \\
\midrule

\multirow{4}{*}{\bf Overall (4,530 samples)} 
& Comments    & 30.50 & 28 & 3 & 800 & 2.82 & 240 & 75.83 & 138179 \\
& Target      & 1.48  & 1 & 1 & 44 & 1.00 & 24 & 0.33  & 6709 \\
& Intent      & 14.65 & 14 & 5 & 40 & 1.02 & 34 & 1.41  & 66305 \\
& Implication & 15.15 & 15 & 5 & 48 & 1.01 & 40 & 0.82  & 68584 \\
\bottomrule
\end{tabular}%
}
\caption{Quantitative summary of the HateMirage dataset, reporting key statistics for the complete dataset and its train and test splits. Metrics include text length, sentence structure, and multi-sentence prevalence, providing an overview of the linguistic complexity and scale of the dataset.}
\label{tab:quant_summary}
\end{center}
\end{table*}


\section{Benchmarking, Experimental Results and Analysis} \label{exp_results}
We evaluate a diverse set of language models, ranging from small to large, to assess their ability to generate structured explanations for faux hate comments. For small language models (SLMs), we include LLaMA variants~\cite{grattafiori2024llama} such as LLaMA-3.2-1B, LLaMA-3.2-3B, Qwen variants~\cite{yang2025qwen3} such as Qwen-2.5-1.5B, Qwen-2.5-3B, and Microsoft Phi model~\cite{abdin2024phi} such as Phi-3-128k-3B. For large language models (LLMs), we consider Mistral-v0.3-7B \cite{jiang2023clip} and LLaMA-3.1-8B-Instruct. These models were chosen to represent diverse architectures, parameter scales, and training paradigms, allowing a systematic comparison of their reasoning and explanation generation capabilities.


The task is formulated as a structured explanation generation problem, where the model is prompted to extract and articulate the Target, Intent, and Implication fields for a given comment. In the zero-shot setting, models are provided only with the comment and a clear instruction describing each field’s meaning and expected format. No fine-tuning or supervised adaptation is performed. We use greedy decoding for all generations to ensure deterministic and reproducible outputs, avoiding variability introduced by sampling-based methods.

We evaluate models in two complementary settings. In the zero-shot setting, we examine the models’ internal reasoning abilities by relying solely on their pretrained knowledge and instructions, without any external context. In the RAG setting, we incorporate additional grounded knowledge, as faux hate often involves specific misinformation narratives that require external verification. We constructed a vector database using FAISS, populated with documents sourced from fact-checking platforms. All documents were preprocessed to remove noise and split into passage-level chunks suitable for retrieval.

For each comment, we performed a semantic search using cosine similarity to retrieve the top five most relevant evidence documents. These retrieved documents were prepended as context in the input prompt, allowing the models to generate explanations of Target, Intent, and Implication grounded in verified factual information. To ensure fully deterministic and reproducible outputs across all benchmarked models, we used greedy decoding with temperature = 0 and do\_sample = False for all inference tasks. This setup guarantees that results can be replicated consistently, supporting reproducibility and benchmarking across different model configurations.

For evaluation, we compare the generated Target, Intent, and Implication explanations against the gold annotations using multiple complementary metrics: Sentence-BERT similarity \cite{reimers2019sentence} and ROUGE-L F1 \cite{lin2004rouge}. Sentence-BERT similarity captures semantic alignment, while ROUGE-L F1 measures lexical and structural overlap. Together, this combination provides a balanced assessment of both meaning and surface form across all explanation fields.

\begin{table*}[!ht]
\begin{center}
\resizebox{0.8\textwidth}{!}{%
\begin{tabular}{ l l cc cc cc }
\toprule
\multirow{2}{*}{\textbf{Model}} & & \multicolumn{2}{c}{\textbf{Target}} & \multicolumn{2}{c}{\textbf{Intent}} & \multicolumn{2}{c}{\textbf{Implication}} \\ \cmidrule(lr){3-4} \cmidrule(lr){5-6} \cmidrule(lr){7-8}
 & & \textbf{SBERT} & \textbf{R-L (F1)}  & \textbf{SBERT} & \textbf{R-L (F1)}  & \textbf{SBERT} & \textbf{R-L (F1)} \\
\midrule
LLaMA-3.2-1B-Instruct & \multirow{7}{*}{\bf Zero-shot} & 50.19 & 30.91 & 57.99 & 26.25 & 47.04 & 12.79 \\
LLaMA-3.2-3B-Instruct & & 53.43 & 34.62 & 57.67 & 26.81 & 47.76 & 16.36 \\
LLaMA-3.1-8B-Instruct & & 48.29 & 21.39 & 56.68 & 27.95 & 51.24 & 15.62 \\
Qwen-2.5-1.5B & & 53.57 & 35.93 & 57.86 & 25.56 & 43.41 & 14.45 \\
Qwen-2.5-3B-Instruct & & 52.91 & 36.13 & 59.96 & 19.06 & 53.09 & 13.75 \\
Phi-3-128k-3b-Instruct & & \textbf{65.55} & \textbf{50.36} & 61.11$^\dagger$ &\textbf{29.52}& 50.39 & 17.27$^\dagger$ \\
Mistral-v0.3-7B-Instruct & & 59.81 & 40.38 & 60.75 & 27.74 & \textbf{55.64} & \textbf{17.39} \\
\midrule
LLaMA-3.2-1B-Instruct & \multirow{7}{*}{\bf RAG} & 52.43 & 30.92 & 57.77 & 25.14 & 47.38 & 13.20 \\
LLaMA-3.2-3B-Instruct & & 49.55 & 28.50 & 55.98 & 25.80 & 48.74 & 16.42 \\
LLaMA-3.1-8B-Instruct & & 40.37 & 14.80 & 59.00 & 26.70 & 52.17 & 14.67 \\
Qwen-2.5-1.5B & & 62.18 & 47.33 & 55.58 & 21.82 & 44.31 & 13.71 \\
Qwen-2.5-3B-Instruct & & 53.86 & 37.97 & 61.03 & 18.15 & 51.31 & 12.82 \\
Phi-3-128k-3b-Instruct & & 63.65$^\dagger$ & 47.81$^\dagger$ & \textbf{62.03} & 26.90$^\dagger$ & 51.20 & 16.25 \\
Mistral-v0.3-7B-Instruct & & 63.06 & 44.27 & 60.03 & 26.38 & 53.68$^\dagger$ & 16.17 \\
\hline
\end{tabular}%
}
\caption{Comparison of LLM performance across Zero-Shot and RAG-based settings. Highest results are highlighted in \textbf{bold}, and second-best results are marked with $^\dagger$. (Metrics reported in \%)}
\label{tab:results}
\end{center}
\end{table*}

From \autoref{tab:results}, it is evident that model performance varies significantly across model size, architecture, and grounding strategy. In the Zero-Shot setting, the Phi-3-128k-3B-Instruct model clearly dominates in Target identification, achieving the highest SBERT Similarity (65.55\%) and ROUGE-L F1 (50.36\%). For the Intent component, Phi-3 again performs competitively, securing the second-best SBERT Similarity (61.11\%) and the best ROUGE-L F1 (29.52\%), indicating strong capability in capturing the author's underlying motive. The Implication dimension, being the most abstract and context-dependent, is best handled by Mistral-v0.3-7B-Instruct, which leads both metrics (SBERT = 55.64\%, ROUGE-L F1 = 17.39\%). These observations suggest that while larger models exhibit stronger inferential reasoning, smaller models like Phi-3 demonstrate exceptional generalization in identifying factual and role-focused elements of faux hate explanations.

Under the RAG-based setting, performance trends remain consistent, with notable improvements in grounding-sensitive components. The Phi-3-128k-3B-Instruct model retains strong performance, achieving the second-best Target results (SBERT = 63.65\%, ROUGE-L F1 = 47.81\%), the highest Intent similarity (SBERT = 62.03\%), and the second-best Intent ROUGE-L (26.90\%). For Implication, Mistral-v0.3-7B-Instruct again performs robustly (SBERT = 53.68\%, ROUGE-L F1 = 16.17\%), reflecting its capacity to reason about broader societal effects when supported by retrieval context.

Overall, while other models such as Qwen-2.5 and LLaMA-3 variants exhibit moderate yet balanced results, the Phi-3 model consistently shows strong performance across multiple dimensions. One possible explanation is that its training mixture includes substantial synthetic and reasoning-focused data, which may encourage structured and instruction-aligned outputs that resemble the format of our references.

At the same time, similarity to GPT-4 may also play a role. Prior documentation for the Phi family indicates influence from GPT-4 class teacher systems, and because our reference explanations are themselves generated by GPT-4, automatic metrics such as ROUGE-L can partly reward overlap in phrasing or discourse style in addition to the underlying quality of reasoning. Therefore, the observed advantage should be interpreted carefully. The results may reflect a combination of genuine interpretive capability and stylistic alignment with the reference generator.

All experiments are conducted using open-source models, with inference run on a cluster of four NVIDIA GPUs, each with 15GB of memory, totaling 60GB. Both small and large language models are evaluated under identical experimental conditions to ensure a fair comparison.

\subsection{Error Analysis}
To better understand the limitations of the benchmarked models in the Faux Hate Explanation task, we conducted a qualitative error analysis on selected examples across the three explanation dimensions: Target, Intent, and Implication (see \autoref{tab:error_analysis_blocks}). For Target prediction, qualitative inspection suggests that in several instances the system relies on surface-level cues such as prominent nouns or early-mentioned entities, which can lead to missing the true focus of the comment. For example, in sarcastic or indirect messages, the model may assign the label \texttt{<organization>} even when the criticism is directed toward a journalist or media actor.

For Intent, we observe multiple cases where the model produces generic or templated explanations such as “spreading hateful or fake narratives,” which may overlook more nuanced motives including sarcasm, racialized mockery, or political positioning. Implication appears to be the most challenging dimension. In several reviewed examples, models generate broad or underspecified outcomes, for instance “promoting general negativity or unrest,” while failing to capture more specific social consequences such as scapegoating, perceived institutional bias, or the reinforcement of double standards. This observation is consistent with our human evaluation results, where Implication also showed comparatively lower agreement than Target and Intent.

Overall, these patterns indicate a tendency toward overgeneralization and difficulty in capturing pragmatic and context-dependent meaning, underscoring the inherent complexity of multi-dimensional reasoning in Faux Hate Explanation.

\begin{table*}[!ht]
\begin{center}
\resizebox{\textwidth}{!}{%
\begin{tabular}{l p{7cm} p{6cm} p{6cm}}
\toprule
& \textbf{Comment} & \textbf{Ground Truth} & \textbf{Model Prediction} \\
\midrule

\multirow{6}{*}{\rotatebox{90}{\textbf{Target}}} & \textit{During the time of Corona, you also actively spread information about the \texttt{<organization>}. Have some shame for you were metaphorically ``hit with a shoe on your face.'' Please report the news with a bit of discretion.} & Media / journalist & \texttt{<organization>} \\ \cmidrule{2-4}

& \textit{Hey Brother, \texttt{<organization>} is spreading the coronavirus here because you're not paying attention to it.} & \texttt{<organization>} & Brother \\ \midrule

\multirow{6}{*}{\rotatebox{90}{\textbf{Intent}}} & \textit{\texttt{<@user>} Were the people from the group in Markaz there to catch corona?} & To question the motives of the \texttt{<organization>} gathering and imply reckless behavior. & The commenter is questioning the intentions of people in Markaz, suggesting they might be there to spread misinformation about the COVID-19 pandemic. \\ \cmidrule{2-4}

& \textit{I still advocate calling it the ``kung-flu''. And label anyone suspected of being infected as ``Flu Kang''.} & To use racially insensitive language to mock and stigmatize people, especially Asians, in relation to COVID-19. & To spread hateful and fake narratives about the flu. \\ \midrule

\multirow{6}{*}{\rotatebox{90}{\textbf{Implication}}} & \textit{\texttt{<@user>} Why.. Did they have corona too? Oh, so you all have been spreading this disease since then... We've been blaming the \texttt{<organization>} for no reason... This has been going on traditionally.} & It implies that the criticism directed toward \texttt{<organization>} was unwarranted and highlights bias or unfair treatment. & The comment could be interpreted as promoting general negativity or unrest, possibly encouraging harmful behavior. \\ \cmidrule{2-4}

& \textit{If \texttt{<organization>} is responsible, then Namaste Trump is more responsible. \texttt{<person>} had given a warning on February 11 regarding Corona. He was mocked then. Who will you hold responsible for Namaste Trump? Trump or our PM? \#shameOnRupani} & This criticism aims to highlight perceived double standards and questions leadership decisions in handling the COVID-19 situation. & The comment could be interpreted as drawing a general comparison between different parties, possibly implying blame or criticism. \\
\bottomrule
\end{tabular}%
}
\caption{Error analysis showing model predictions for Target, Intent, and Implication. Placeholders such as \texttt{<organization>} and \texttt{<user>} are used to anonymize entities and generalize examples.}
\label{tab:error_analysis_blocks}
\end{center}
\end{table*}

\subsection{Human Evaluation} \label{human_eval}
The HateMirage dataset was generated using the GPT-4 model augmented with contextual information through Retrieval-Augmented Generation (RAG). To assess the quality of GPT-4-generated explanations in the HateMirage dataset, we first compared them against human-written explanations. For this, 500 randomly selected records were independently annotated by trained human annotators, who produced their own explanations for Target, Intent, and Implication. The GPT-4 outputs were then quantitatively evaluated using SBERT cosine similarity and ROUGE-L F1 scores against the human-written references as shown in \autoref{tab:human_eval_scores}. SBERT primarily reflects semantic agreement, whereas ROUGE-L captures lexical and phrasal overlap. Because multiple valid explanations can differ in wording while preserving meaning, it is possible to observe relatively high SBERT values alongside more moderate ROUGE scores. Notably, 208 of 500 records surpassed a 90\% similarity threshold for Target, only 7 for Intent, and none for Implication, highlighting that GPT-4 reliably identifies entities and general motives, while reasoning over broader causal consequences and nuanced implications remains challenging. This pattern is consistent with our independent human rating study, where Implication likewise received lower agreement relative to Target and Intent.

\begin{table}[H]
\begin{center}
{%
\begin{tabular}{lcc}
\toprule
 & \textbf{SBERT} & \textbf{R-L (F1)} \\
\toprule
Target & 0.7479 & 0.6293 \\
Intent & 0.7020 & 0.3458 \\
Implication & 0.5780 & 0.2013 \\
\bottomrule
\end{tabular}%
}
\caption{Human evaluation results comparing GPT-4 generated explanations with human-written explanations for 500 randomly sampled records.}
\label{tab:human_eval_scores}
\end{center}
\end{table}

\begin{figure}[!ht]
    \centering
    \includegraphics[width=0.45\textwidth]{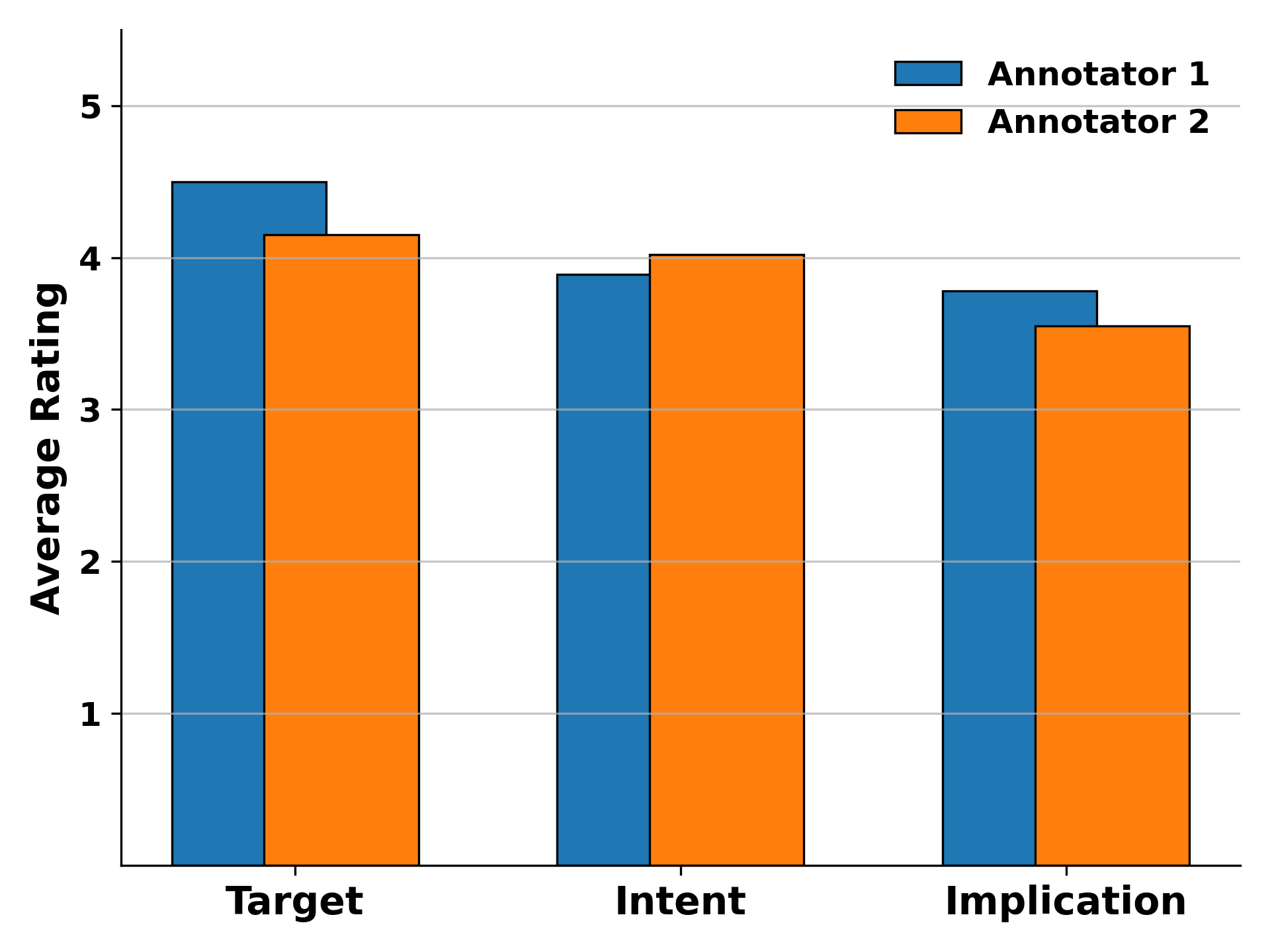}
    \caption{Comparison of average ratings from two annotators for GPT-4-generated explanations across Target, Intent, and Implication.}
    \label{fig:human-eval-rating}
\end{figure}

In addition, we conducted a quantitative human assessment on 100 randomly selected records from the dataset, independently annotated by two human annotators. For each record, annotators rated the GPT-4-generated explanations for Target, Intent, and Implication on a scale from 1 (poor) to 5 (excellent). As shown in \autoref{fig:human-eval-rating}, the majority of ratings for Target fall between 4 and 5, Intent is close to 4, and Implication ranges between 3 and 4. These results indicate that while GPT-4 effectively captures entities and general motives, capturing broader causal consequences and nuanced societal implications remains more difficult, emphasizing the inherent challenge of reasoning over downstream effects in faux hate content.

\section{Conclusion and Future Work} \label{conclusion}
In this work, we introduced \textit{HateMirage}, a novel dataset designed to advance the study of subtle and deceptive forms of harmful online content. Unlike traditional hate speech or misinformation corpora, HateMirage focused exclusively on Faux Hate, the content where hateful narratives emerge indirectly through misinformation or deceptive framing. Each comment was annotated with structured, multi-dimensional explanations capturing \textit{Target}, \textit{Intent}, and \textit{Implication}, thereby uncovering the implicit layers of meaning that are often missed by surface-level analysis. By providing these rich explanatory annotations, HateMirage aimed to support both research and moderation efforts by making the underlying harmful reasoning more transparent and interpretable, even to non-expert users. We further benchmarked a range of language models on explanation generation tasks, establishing initial baselines and revealing key challenges in structured reasoning over misinformation-driven hate. These results highlighted significant gaps between current model capabilities and the complexity of reasoning required to interpret Faux Hate.

In future work, we plan to extend HateMirage to the multimodal domain by incorporating memes and visual narratives, enabling the study of how text-image interactions shape deceptive hateful discourse across different social media platforms. Additionally, we aim to explore automated methods for generating and evaluating multi-dimensional explanations, including explanation faithfulness metrics and model-based evaluators that move beyond lexical similarity to better assess reasoning consistency and grounding, paving the way for more interpretable and robust moderation systems.

\section{Limitations} \label{limitations}
While the HateMirage dataset is a unique resource providing structured, multi-dimensional explanations for Faux Hate comments, certain limitations must be acknowledged.

The primary limitation lies in the synthetic nature of the explanations. The annotations were generated using GPT-4 with RAG. This approach was deliberately chosen to minimize the exposure of human annotators to highly distressing and toxic content, as Faux Hate can be psychologically disturbing; accordingly, human verification was conducted on a small, random subset of 500 records (approximately 11\% of the total dataset). Despite human-in-the-loop validation, these model-generated explanations may lack the full subtlety and emotional variability of purely human interpretations. This is particularly notable in the abstract Implication field, which proved most difficult for the model to infer accurately against human-written gold standards. Furthermore, the corpus is derived specifically from YouTube comments on international English news channels, meaning it may not fully represent the vernacular and platform-dependent characteristics of Faux Hate spread across all social media environments. Despite these limitations, HateMirage provides a valuable foundation for studying Faux Hate and developing reasoning and moderation-oriented systems.

\section{Ethics Statement}
The HateMirage dataset is constructed from publicly available YouTube comments posted under international news videos. Data collection strictly followed YouTube’s Terms of Service\footnote{\url{https://www.youtube.com/t/terms}} and Privacy Policy\footnote{\url{https://policies.google.com/privacy}}. Only publicly accessible comments were included; no private or restricted data were used. Given the sensitive nature of the content, strict anonymization procedures were applied: all personally identifiable information (PII), including usernames, profile links, and timestamps, was not collected, and named entities were masked where appropriate to minimize risks of re-identification and to avoid amplifying harmful narratives. The dataset is intended solely for academic research on explainable NLP, misinformation, and hate speech. Public redistribution is strictly prohibited. Access will be provided under a controlled, research-only license requiring requesters to agree to responsible-use terms, and any form of commercial use or generative model training outside a research setting is prohibited.

\paragraph{Broader Impact:} 
By focusing on explanatory reasoning rather than simple classification, this work aims to enable more transparent and accountable moderation systems and support research on the interplay between misinformation and hateful discourse. Structured explanations can help moderation teams, fact-checkers, and researchers better understand why content is harmful, thereby promoting safer online spaces and more interpretable AI systems. However, potential risks remain: malicious actors could misuse structured explanations to craft more sophisticated hateful narratives or target vulnerable groups. To mitigate this, we restrict dataset access, anonymize all entities, and prohibit redistribution or use in generative settings that could amplify harm. Overall, this work seeks to advance responsible scientific analysis of nuanced online harm while safeguarding user privacy, minimizing misuse, and reflecting critically on the societal implications of explainable NLP systems in sensitive domains.




\section*{Acknowledgement}
M. S. Akhtar acknowledges the partial support of Infosys Center of AI (CAI) at IIIT Delhi.

\section{Bibliographical References}\label{sec:reference}

\bibliographystyle{lrec2026-natbib}
\bibliography{lrec2026-example}


\end{document}